\documentclass{article}

\usepackage{times}
\usepackage{graphicx} 
\usepackage{subcaption} 

\usepackage{natbib}

\usepackage{algorithm}
\usepackage{algorithmic}

\usepackage{hyperref}

\usepackage[accepted]{icml2016}

\usepackage{relsize} 
\usepackage{amsmath} 
\usepackage{amssymb}

\usepackage{empheq} 

\newcommand{\E}{\mathop{\mathlarger{\mathbb{E}}}}
\newcommand{\muhat}{\hat{\mu}}

\newcommand{\todo}[1]{}
\renewcommand{\todo}[1]{{\color{red} TODO: {#1}}}
\newcommand{\question}[1]{}
\renewcommand{\question}[1]{{\color{blue} QUESTION: {#1}}}

\makeatletter
\newcommand*\dashline{\rotatebox[origin=c]{90}{$\dabar@\dabar@\dabar@$}}
\makeatother

\newtheorem{theorem}{Theorem}[section]

\newtheorem{lemma}[theorem]{Lemma}

\icmltitlerunning{Ignoring the Distracting Features}

\begin{document} 

\twocolumn[
\icmltitle{Ignoring Distractors in the Absence of Labels: Optimal Linear Projection to Remove False Positives During Anomaly Detection}

\icmlauthor{Allison Del Giorno}{adelgior@cs.cmu.edu}
\icmladdress{Carnegie Mellon University,
            5000 Forbes Ave, Pittsburgh, PA 15213 USA}
\icmlauthor{J. Andrew Bagnell}{dbagnell@ri.cmu.edu}
\icmladdress{Carnegie Mellon University,
            5000 Forbes Ave, Pittsburgh, PA 15213 USA}
\icmlauthor{Martial Hebert}{hebert@ri.cmu.edu}
\icmladdress{Carnegie Mellon University,
            5000 Forbes Ave, Pittsburgh, PA 15213 USA}

\icmlkeywords{anomaly detection, feature selection}

\vskip 0.3in
]
\begin{abstract}
In the anomaly detection setting, the native feature embedding can be a crucial source of bias.  We present a technique, Feature Omission using Context in Unsupervised Settings (FOCUS) to learn a feature mapping that is invariant to changes exemplified in training sets while retaining as much descriptive power as possible.  While this method could apply to many unsupervised settings, we focus on applications in anomaly detection, where little task-labeled data is available. Our algorithm requires only non-anomalous sets of data, and does not require that the contexts in the training sets match the context of the test set.  By maximizing within-set variance and minimizing between-set variance, we are able to identify and remove \textit{distracting features} while retaining fidelity to the descriptiveness needed at test time.  In the linear case, our formulation reduces to a generalized eigenvalue problem that can be solved quickly and applied to test sets outside the context of the training sets.  This technique allows us to align technical definitions of anomaly detection with human definitions through appropriate mappings of the feature space.  We demonstrate that this method is able to remove uninformative parts of the feature space for the anomaly detection setting.


\end{abstract}


\begin{figure*}[t]
\centering
\begin{subfigure}{.33\linewidth}
  \centering
  \includegraphics[width=.95\linewidth]{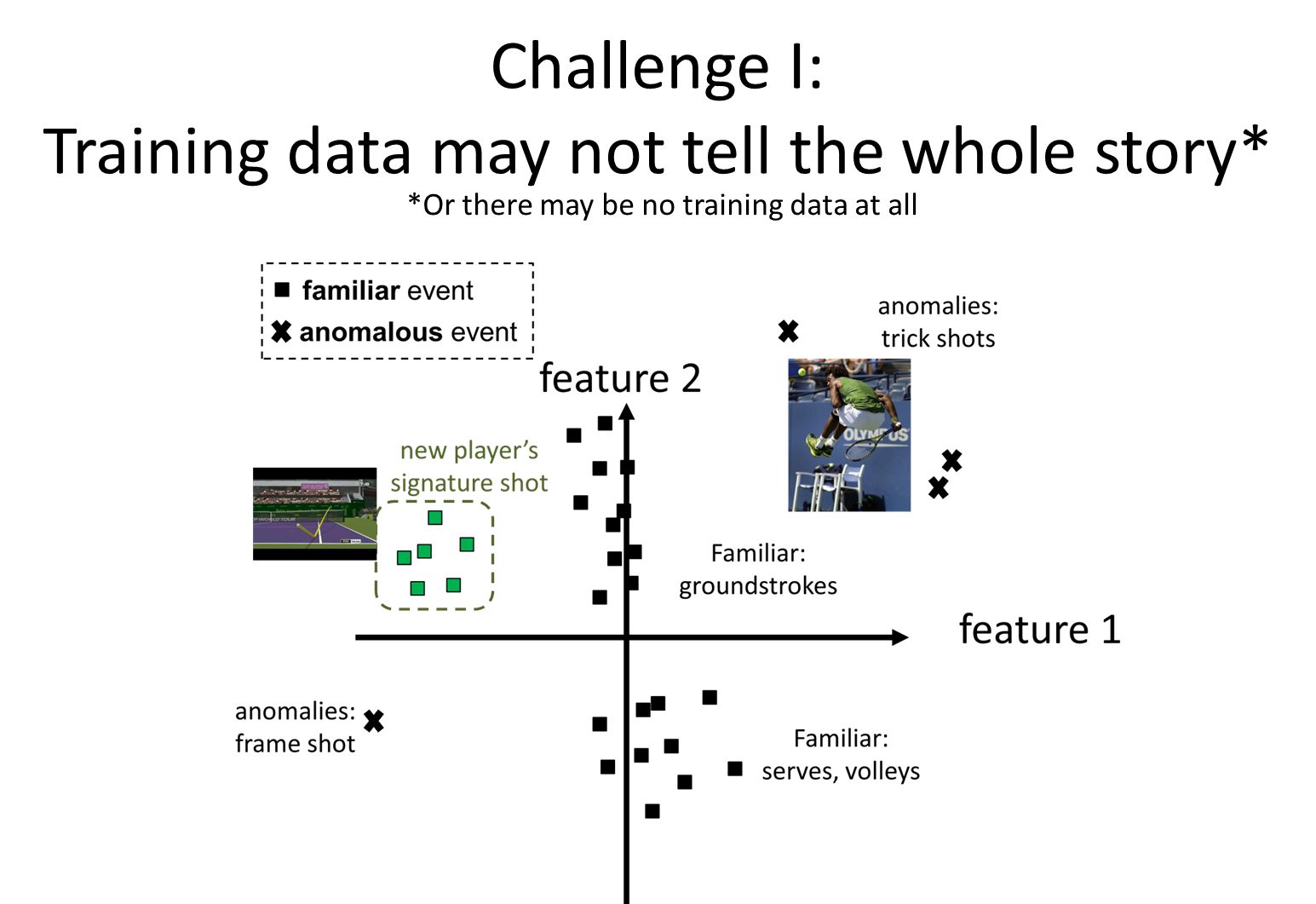}
  \label{fig:challenge1}
\end{subfigure}%
\begin{subfigure}{.33\linewidth}
  \centering
  \includegraphics[width=.95\linewidth]{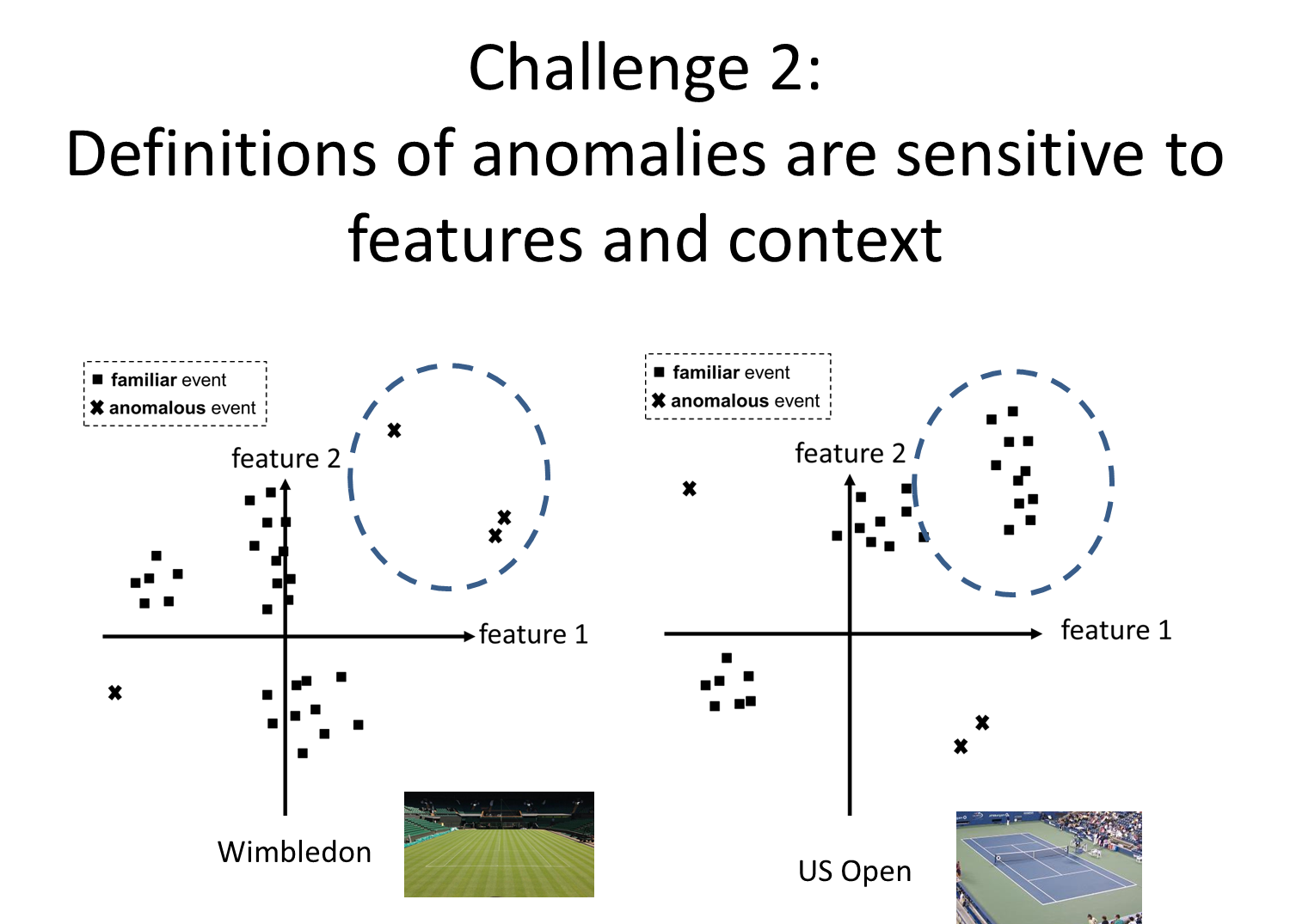}
  \label{fig:challenge2}
\end{subfigure}
\begin{subfigure}{.33\linewidth}
  \centering
  \includegraphics[width=.95\linewidth]{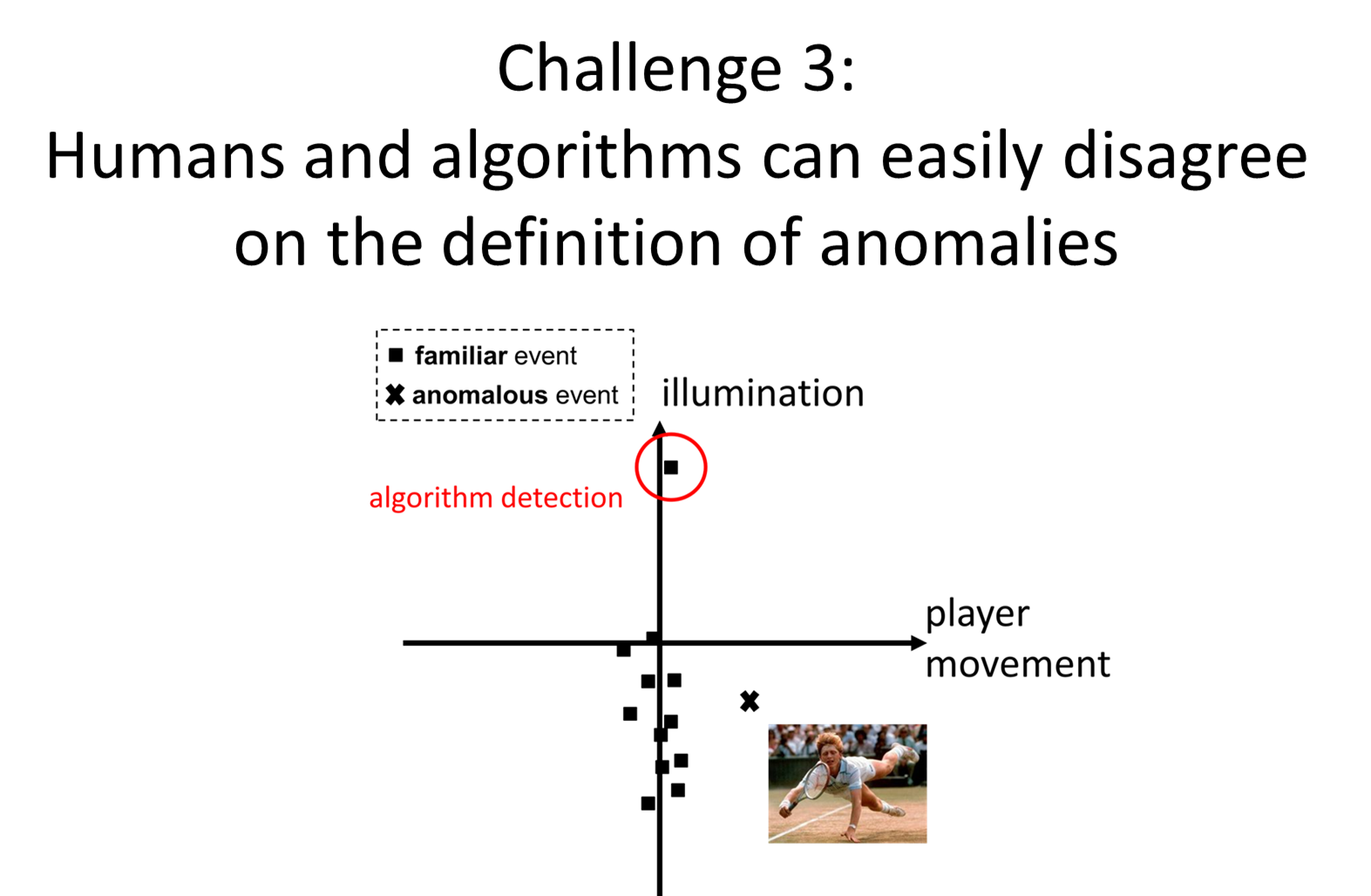}
  \label{fig:challenge3}
\end{subfigure}
\caption{\textbf{Challenges particular to anomaly detection. }
(1) We cannot use labeled data for anomalies -- it is impossible to find a sufficiently representative set; (2) The distribution of the normal, or \textit{familiar} class can sometimes be learned from a set of familiar training examples if the context of the test video is known ahead of time. Most of the time, it must be inferred from common events in the test video itself.
\textbf{Left.} Full training data is rarely available for this problem, and available data can be misleading (e.g. - a player may switch strategies during a match).  The new strategy becomes familiar but looks anomalous in the context of the training data. \textbf{Middle.} The context of a different video may change the definition of an anomaly.  The features that indicated familiar events in one video may indicate anomalies in another video.  \textbf{Right. } Human labels are subjective.  Humans are agnostic to many features that algorithms can easily pick out.  Some type of supervision is required to reduce these errors.
}
\label{fig:challenges}
\end{figure*}

\section{Introduction}
\textbf{Anomaly detection relies heavily on its feature embedding.} Anomaly detection algorithms are at the mercy of the feature embedding in which they operate.  The identity of anomalies are context-dependent: the answer for a given point in feature space depends on the distribution of points around it.  For instance, a tennis player hitting a 120mph serve will seem anomalous in the midst of a novice match, but it is commonplace in a professional match.  Because the ground truth changes from one set of points to another, effective algorithms must adapt to the test set, sometimes exclusively relying on the distribution of points in the test set to define context.

In order for anomaly detection to work, (1) anomalous points must be distinguishable from familiar points, and (2) familiar points must be less distinguishable from each other.  Condition 1 is met by any sufficiently descriptive feature space.  However, in most feature spaces, condition 2 is rarely met: often, variations that we will call \textit{distracting features}, such as camera shake and illumination changes, make familiar points seem more easily distinguished from each other, and the algorithm flags false positives that humans find irrelevant.
As a result, anomaly detection algorithms have a tendency to latch on to parts of the feature space that are unimportant.  These variations in test sets are a large source of false positives in anomaly detection methods, and the ambiguity of which variations humans do and do not care about makes it difficult to overcome this issue.

Some methods learn context-specific invariances from a set of familiar data that must match the familiar distribution in the test data.  With enough data and examples in each context, a model-based technique could fully represent the space of `normal' events and mark anomalies as points outside of that distribution.  In practice, however, there is often a lack of supervised data to fully describe what is familiar (or what a good metric for deviation from the model looks like).  In addition, the model learned in one context cannot be applied to another context, so this requires a new set of labeled familiar data for each context.  





\textbf{Supervised learning overcomes this problem with task-driven labels.}  Recent developments in feature learning have led to large gains in supervised computer vision tasks by jointly learning a feature mapping with a labeling task.  Neural networks use nonlinear hierarchies to generate an encoding optimized over millions of datapoints for a given task.  The invariances learned in those networks have been well-studied~\cite{invariancenns}.  One of the most established techniques is to compute the optimal linear mapping for classification analytically using Linear Discriminant Analysis (LDA)~\cite{lda}.  A supervised learning algorithm's objective is to map instances within a class close to each other and ensure different classes are separable.  With enough training data matching the testing distribution, all other information that does not meet this objective on the training set can be safely removed.  This means that during tasks such as object recognition or segmentation, \textit{distracting features} are removed as a byproduct of the learning task: the algorithm's objective is met as it learns to keep predictions consistent across invariances found in large amounts of human-labeled data.  For instance, cars, trees, and people retain their labels as lighting sources or camera angles change.  

\textbf{Anomaly detection cannot leverage the same techniques in the absence of task-labeled data.} However, the anomaly detection setting is unsupervised: by construction, there are not labels of the anomaly class, and often there is not even a representative set of familiars that cover the testing context (Figure~\ref{fig:challenges}).  There are no absolute boundaries that can be drawn between familiar and anomalous points for any testing context; the methods must adapt to the testing context.  Without a universal set of labels, it is not possible to jointly learn the task as well as features as supervised methods do.

Unsupervised methods such as Principal Component Analysis (PCA), Locally Linear Embedding (LLE)~\cite{lle}, and Dr.LIM~\cite{drlim} are used to evenly distribute distances between points when we can assume training and testing distributions are similar.

We aim to bring feature learning to anomaly detection in a way that (1) does not require task-based labels and (2) remains conservative enough to avoid impeding context-based methods.  This paper presents one way of learning the invariances for anomaly detection by learning a feature transformation from sets of examples containing only familiar events.  None of these examples need to be drawn from the same context as the test set, and no examples of anomalies need to be provided.  This allows us to still use heavily context-driven algorithms for anomaly detection, leveraging existing techniques while boosting performance by addressing a key challenge of learning feature embeddings for anomaly detection (Figure~\ref{fig:challenges}).

\section{Problem Formulation}

\begin{figure*}[t]
\centering
\begin{subfigure}{.35\linewidth}
  \centering
  \includegraphics[width=.95\linewidth]{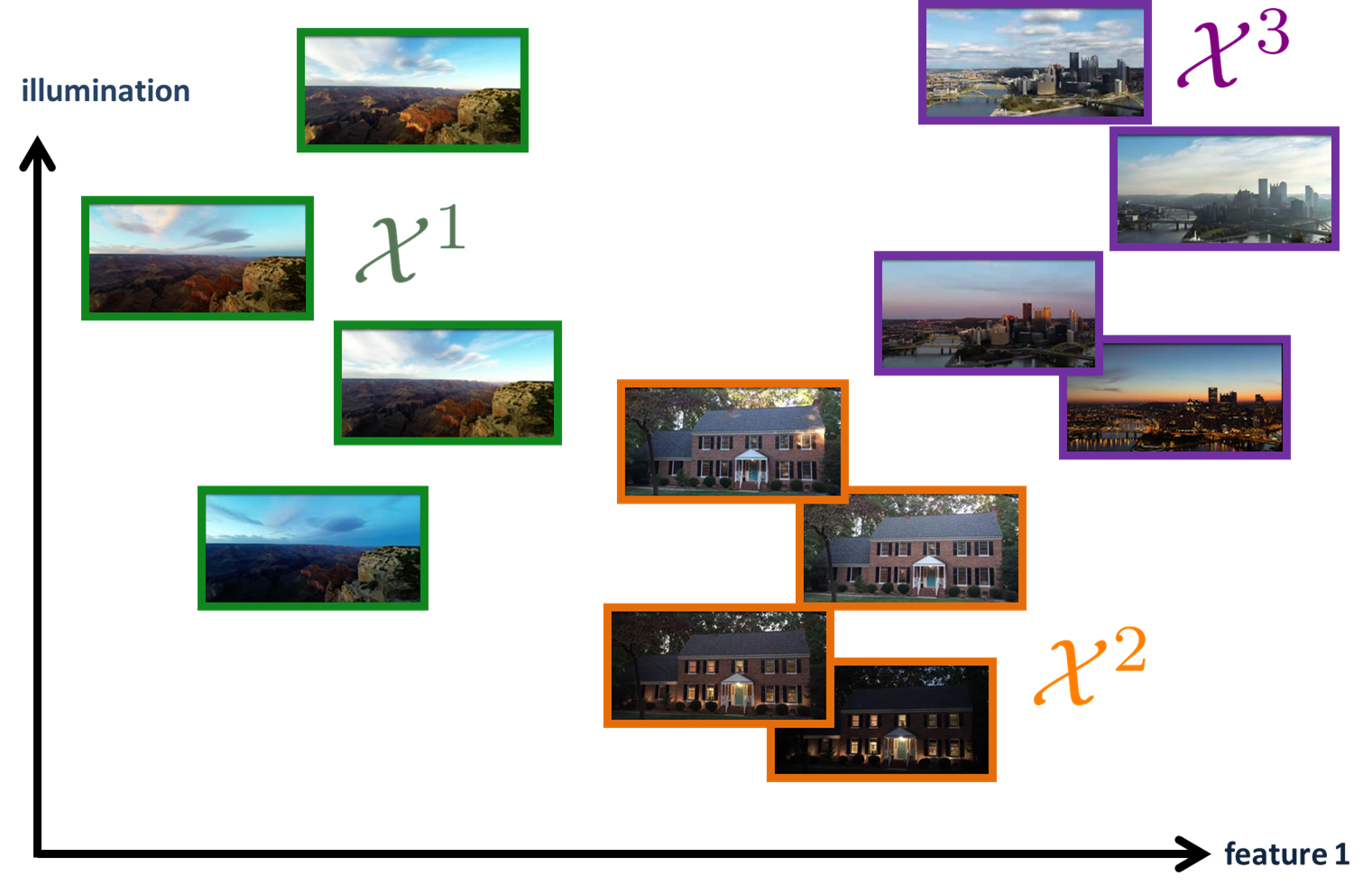}
  \caption{Provided training sets of non-anomalous data}
  \label{fig:traina}
\end{subfigure}%
\hspace{.1\linewidth}
\begin{subfigure}{.35\linewidth}
  \centering
  \includegraphics[width=.95\linewidth]{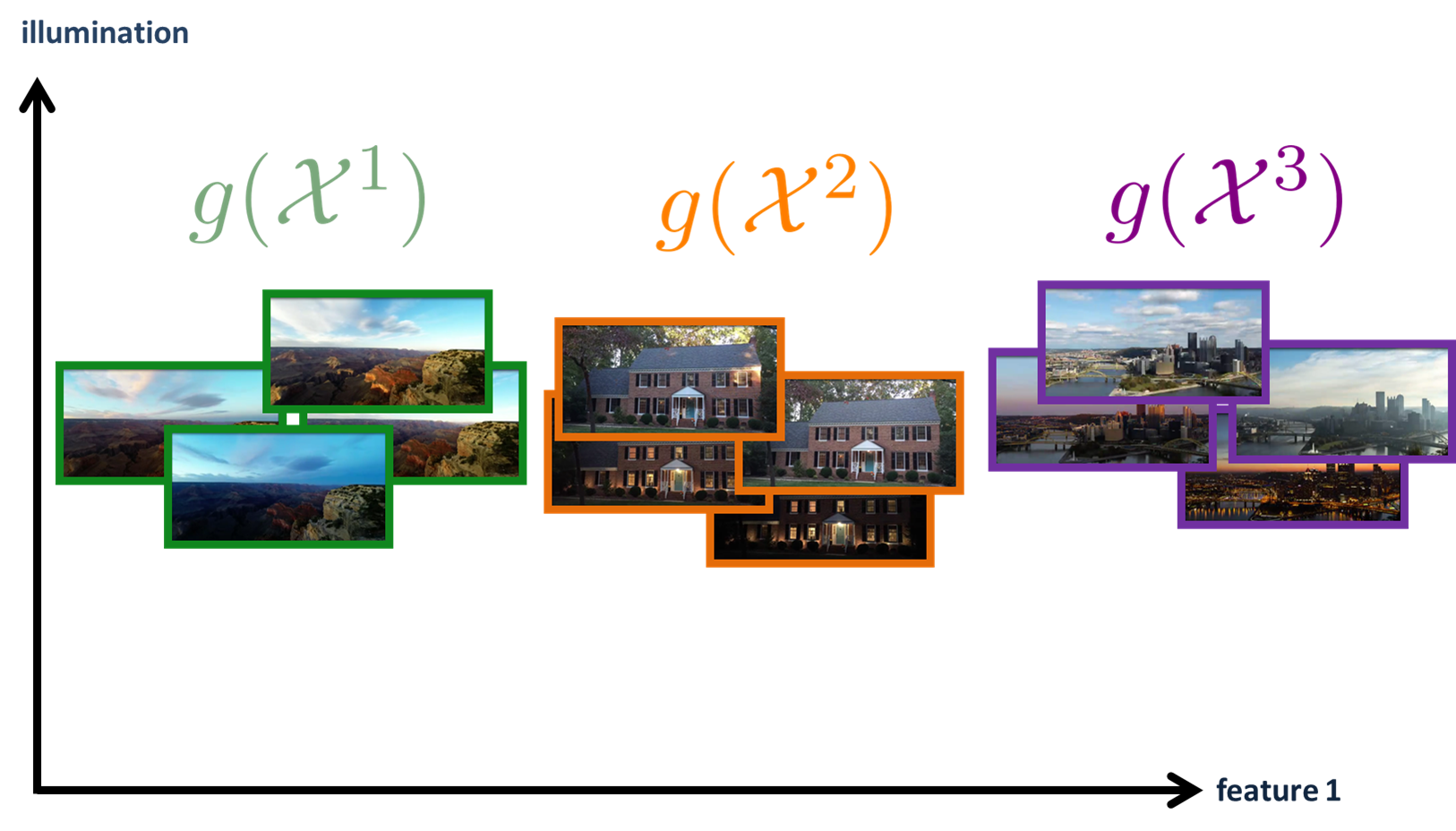}
  \caption{Feature mapping is learned from training sets}
  \label{fig:trainb}
\end{subfigure}
 \vskip\baselineskip
 \begin{subfigure}{.35\linewidth}
  \centering
  \includegraphics[width=.95\linewidth]{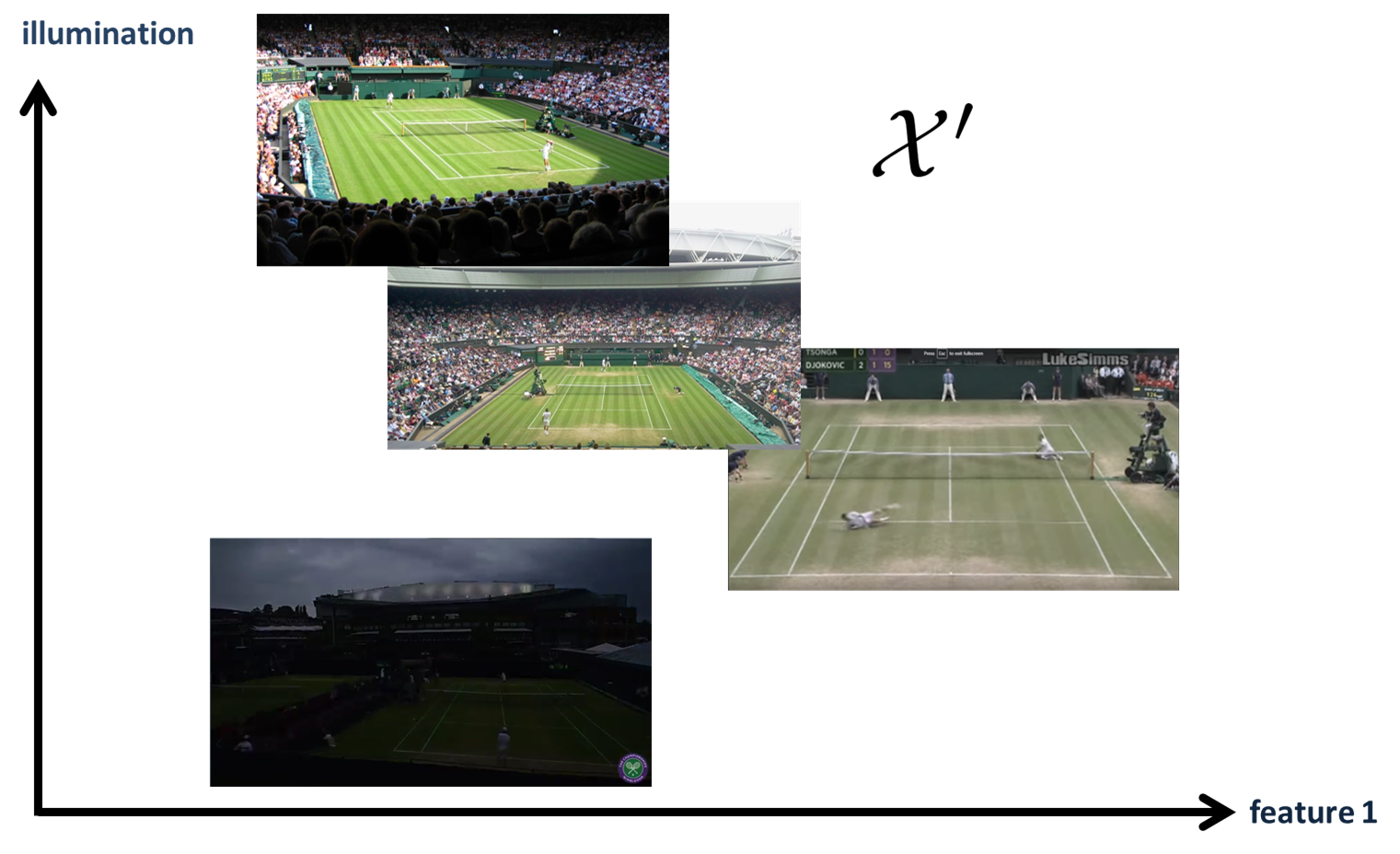}
  \caption{Test set}
  \label{fig:testa}
\end{subfigure}
\hspace{.1\linewidth}
\begin{subfigure}{.35\linewidth}
  \centering
  \includegraphics[width=.95\linewidth]{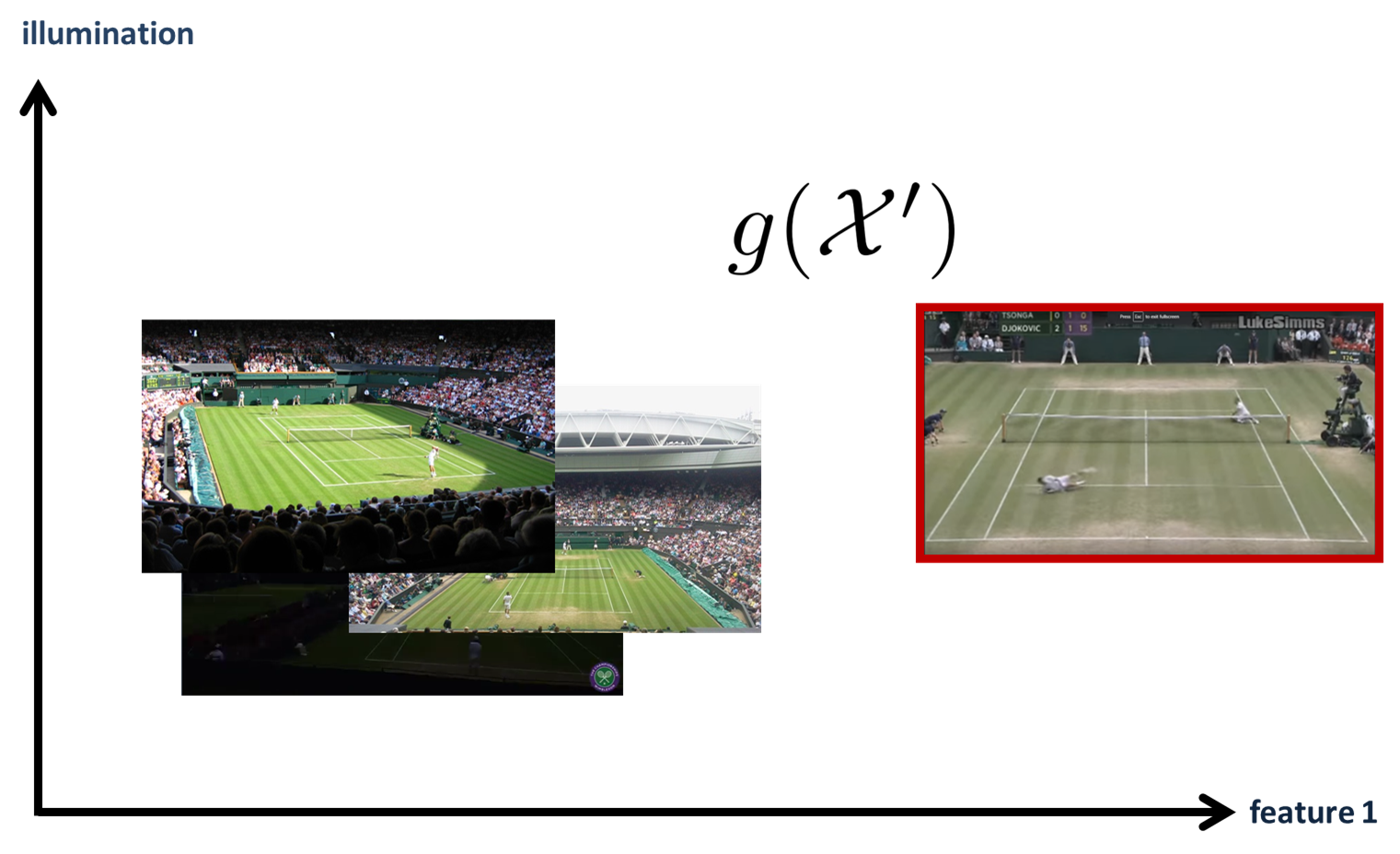}
  \caption{Feature mapping is applied to test set}
  \label{fig:testb}
\end{subfigure}
\caption{\textbf{Problem Setup and Objective. } Groups of non-anomalous data are provided to the algorithm.  Points within each set are from the same context; each set contains different context.  A feature mapping $g(x)$ is learned, maximimizing the intraset variance and minimizing the interset variance.  The complementary mapping $g*(x)$ is then applied to a test set in any context (even one different from that of the training sets), making human-relevant anomalies easier to find in the new, informative feature space.  For ease of illustration, the feature mapping is axis-aligned in this case.  This need not be the case in general.
}
\label{fig:problemoverview}
\end{figure*}

\textbf{The test set.}  We have been given a test set $X'$ with $n'$ data points, each of which is a $d$ dimensional feature descriptor.  For instance, we may be given a video of $n'$ frames represented by a $d$-dimensional dense optical flow vector or CNN fc7 encoding for each frame.  Our objective is to compute an anomaly score $a'_i$ for each of the $n'$ data points (frames) in the test set.

\textbf{Training sets.}  Suppose we are also given $M$ training sets that contain only familiar data.  Each training set resides in the same feature space as the test set (optical flow for $M$ videos): $\mathcal{X}^j \in \mathbb{R}^{n_j \times d}$, $j \in {1,..,M}$.

Our objective is to isolate and remove the \textit{distracting features} from the feature space.  Because the training data is known to be normal, these distracting features are represented by functions that allow us to tell points within a set apart.  While the exact metric defining distinguishability varies between algorithms, most rely on some measure of relative or absolute distances between points.  We therefore aim to make it difficult to distinguish possible false positives from each other by identifying directions that contribute to the distance between the familiar points.  Once we have identified functions $g(x)$ that represent these distracting features, we can remove them (by applying a complementary feature mapping $g^*(x)$) and give the remaining features to an anomaly detection algorithm for processing.  We will focus our formal objective on euclidean distances in the native feature space, but the feature omission should provide benefits for algorithms that rely on the discriminability of points according to linear metrics, such as sparse generative models~\cite{lu2013abnormal}, density ratios and linear classifiers~\cite{delgiorno2016discriminative}, and standard one-class SVMs.

\subsection{General formulation}

We are given a set of $M$ training sets:
$$\mathcal{X}=\{\mathcal{X}^1,...,\mathcal{X}^M\}$$
where set $\mathcal{X}^m$ contains $n_m$ points.

\subsubsection{Invariance. }
We would like to find directions that are responsible for the within-set variation.  One could consider optimizing this objective directly by maximizing:
\begin{empheq}{align}~\label{eqn:Lwithin}
    L^\text{within}_{g\in\mathcal{G}}(\mathcal{X}) = & \sum_m \sum_{\substack{x,y\in \mathcal{X}^m}} ||g(x)-g(y)||_2^2
\end{empheq} for some set of functions $\mathcal{G}$.
Because each training set is known to be normal relative to itself, functions that maximize this term represent features that would distract an anomaly detection algorithm if it were run on each set of data individually.  If the set of functions $\mathcal{G}$ is finite, this objective can simply be evaluated for each function and we will choose the function that achieves the maximum.  In our setting, we consider $\mathcal{G}$ to be the continuous set of linear functions e.g. - the set of all linear projections ${g: g_i(x)=w_i^Tx}$.  For a single linear projection $g(x)=w^Tx$, $L^\text{within}_g$ reduces to:
\begin{empheq}{align}~\label{eqn:Lwithin_linear}
    L^\text{within}_{w}(\mathcal{X}) = & 2w^TC^\text{within} w
\end{empheq} where $C^\text{within}$ represents the weighted sum of covariance matrices in each training set: 

$$C^\text{within} =\E_m \E_{X|m} (X-\muhat_m)(X-\muhat_m)^T$$
  (see details in Appendix~\ref{section:Cwithin_linear}).

However, simply maximizing Eqn.~\ref{eqn:Lwithin_linear} without a constraint on $g$ leads to a set of unbounded solutions for linear functions (and for most choices of $\mathcal{G}$ with Eqn~\ref{eqn:Lwithin}); i.e. - $\forall x, g(x) = \infty$, and the resulting feature mapping is trivial $g^*(x)=0$.  This is because we have not specified which information to preserve; we therefore need to add a regularizer.

\subsubsection{Regularization. }
\textbf{Option 1: Rank constraint.} One option to preserve information would be to constrain the size of $g$, such as through a rank constraint on $k$ for a set of linear projections.  However, this does not ensure that the descriptive content is preserved.  More importantly, simply optimizing Eqn.~\ref{eqn:Lwithin} with a rank constraint biases our algorithm based on the context of the training set.  Imagine using a few training sets from different sports with lighting changes to encourage illumination invariance, but domain-specific features from that set (horizontal motion of players) have orders of magnitude more variance than the slight effects of the lighting changes.  These domain-specific directions are measured as more distracting than the lighting changes.  Either we carefully choose training sets to ensure all directions of variance can be removed and place a rank constraint on $g$, or we incorporate some notion of what to preserve into our objective.

\textbf{Option 2: Fidelity term.} For anomaly detection, one label-free choice is to preserve the information across all sets by simultaneously maximizing the variance across all training sets:
\begin{empheq}{align}~\label{eqn:Lall}
    L^{\text{all}}_g(\mathcal{X}) = & \frac{1}{n_\text{all}}\sum_{x,y \in \mathcal{X}} ||g(x) - g(y)||_2^2
\end{empheq}

For a single linear projection $g(x)=w^Tx$, this reduces to:
\begin{empheq}{align}~\label{eqn:Lall_linear}
    L^\text{all}_w = 2 w^T C^\text{all} w
\end{empheq} where $C^\text{all}$ is the covariance matrix computed across all points in the training set: 
$$C^\text{all} = \E_m\E_{X|m}\left[(X-\muhat_\text{all})(X-\muhat_\text{all})^T\right]$$
(see details in Appendix~\ref{section:Call_linear}).

We can therefore represent distracting features by maximizing features the anomaly detection algorithm should be invariant to (Eqn~\ref{eqn:Lwithin}) while preserving overall data fidelity (Eqn~\ref{eqn:Lall}):
\begin{empheq}{align}~\label{eqn:geneig_origin}
    \arg\max_g L_g^{\text{within}} - \lambda L_g^\text{all}
\end{empheq}

Optimizing $L^\text{within}_g$ and $L^\text{all}_g$ is feasible for any set of convex loss functions, but can grow expensive for arbitrary functions $g$.  However, for linear functions $g=w^Tx$, the solution can be found more efficiently.  The regularized objective in Eqn.~\ref{eqn:geneig_origin} for linear $g$ reduces to a generalized eigenvalue problem (See proof in Appendix~\ref{section:lossgeneig}).  As a result, the notion of a distracting feature for anomaly detection is formulated as one that optimizes the following objective:
\begin{equation}
\max_{w, ||w||=1} \frac{w^T C^{\text{within}} w}{w^T C^{\text{all}} w}
\label{eqn:ratio_orig}
\end{equation}

In practice, we `cushion' the null space of $C^\text{all}$ by adding a regularizer to the bottom:
\begin{equation}
\max_{w, ||w||=1} \frac{w^T C^{\text{within}} w}{w^T (C^{\text{all}} + \epsilon I) w}
\label{eqn:ratio_with_denom}
\end{equation}

\subsection{Understanding solutions according to the span of covariances}
The following relation between $C^\text{within}$ and $C^\text{all}$ holds: \\
$$C^\text{all} = C^\text{within} + Q$$
where $Q$ is a rank $M-1$ positive semidefinite matrix spanned by the vectors between the means of the training sets (see further details in Appendix~\ref{section:within_between_nullspace_bootstrap}).

The set of eigenvector, eigenvalue solutions $w_i, \lambda_i$ produced by solving the generalized eigenvalue problem in Eqn.~\ref{eqn:ratio_with_denom} fall into three categories (with successive rules):
\begin{itemize}
\item $w_i$ is in the nullspace of $C^\text{within}$ : $\lambda_i=0$
\item else if $w_i$ is in the span of $Q$ : $\lambda_i \in (0,1)$
\item else if $w_i$ is in the nullspace of $Q$ : $\lambda_i=1$
\end{itemize}
Interestingly, at most $M-1$ feature vectors have values between $0$ and $1$; as in Fisher Linear Discriminant Analysis (LDA), the rank of the $Q$ matrix constrains the number of eigenvalues we can uncover.  Unlike LDA, we take a conservative approach: we are only interested in removing directions that seem to be distractors based on the training data.  This means that eigenvectors corresponding to eigenvalues of $0$ should be kept, and the eigenvectors to be removed are those with eigenvalues of $1$ and some subset of those with eigenvalues between $0$ and $1$ (depending on the chosen cutoff).  We will consider the implications of each of the cases below.

\textbf{$\lambda_i=0$: Null space of $C^\text{within}$. }  The null space of $C^\text{within}$ represents the directions in which there is no within-video variation (imagine a feature that is fixed throughout the video, like a date stamp or a dead pixel). Because the features do not vary in these directions, if a point were to vary in this direction, it should be flagged as an anomaly.  Therefore, the null space of $C^{\text{within}}$ is a valid (and important) set of excluded solutions, which we call $\mathcal{W}_{\text{null}}$:

\textbf{$\lambda_i\in(0,1)$: Span of $Q$. }  The span of $Q$ represents the directions in which the sum of squared distances between videos is nonzero; this reduces to the span of vectors formed between centers of training sets.  Because the features do not vary in these directions, if a point were to vary in this direction, it should be flagged as an anomaly.  Therefore, the null space of $C^{\text{within}}$ is a valid (and important) set of excluded solutions, which we call $\mathcal{W}_{\text{null}}$:

\begin{equation} \label{eqn:nullspaceinclude}
\mathcal{W}_{\text{null}} = \{ w : w \in \text{nullspace}(C^\text{within}) \} \\
\end{equation}

\textbf{$\lambda_i=1$: Null space of $Q$. }  Vectors that lie in the span of $C^\text{within}$ and in the null space of $Q$ have corresponding eigenvalues of $1$.  This set of solutions represents directions in which the means of the training videos are the same, though there still exists some variation within the videos.  The number of these eigenvectors decreases as more diverse training videos are given; this makes sense as more information is being presented about what to preserve.

\section{Experiments}

\subsection{A Basic Analytic Example}~\label{section:analytic_example}
Here we present a simple example where the solutions are clear and axis-aligned.

\begin{figure}
\centering
  \centering
  \includegraphics[width=.95\linewidth]{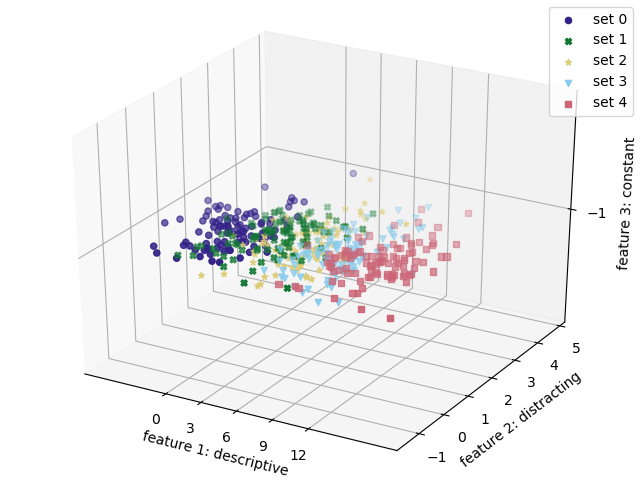}
  \caption{Example training sets drawn from the distributions described in ~\ref{section:analytic_example} with $M=5$ sets.}
  \label{fig:analytic_example_3d}
\end{figure}

Suppose the training sets are drawn from normal distributions as follows (Figure~\ref{fig:analytic_example_3d}):
\begin{align*}
x \in \mathcal{X}^m ~ &\sim  N
\begin{bmatrix}
\begin{pmatrix}
3m\\
1\\
-1
\end{pmatrix}\!\!,&
\begin{pmatrix}
2 & 0 & 0\\
0 & 1 & 0\\
0 & 0 & 0
\end{pmatrix}
\end{bmatrix}
\end{align*}

The first feature dimension is \textit{descriptive}: its value is useful in describing which set a given point belongs to.  The second dimension is \textit{distracting}: it varies within each set and is not helpful in distinguishing between the sets; the training sets should be built to 'highlight' this variation.  The third feature dimension is \textit{constant}.  It does not vary in the training set (but it would be anomalous to see something change in this direction, so we need to keep it).  An example of each of these features in images might be global illumination (distracting), number of people in the scene (descriptive), and a weapon detector (may remain negative throughout the training sets, but important to preserve for anomaly detection in the test sets).

With a sample distribution of $M=10$, $n_m=100 \forall m$, the generalized eigenvalue problem ~\ref{eqn:ratio_with_denom} yields the eigenvectors close to $[e_2, e_1, e_3]$ with corresponding eigenvalues $[0, 0.05, 0.99]$, where $e_d$ is the basis corresponding to feature $d$.  Any reasonable choice of threshold will decide to remove $e_2$, leaving the test set to be projected onto $[e_1, e_3]$.  This keeps both the \textit{descriptive} and \textit{constant} features but successfully removes the distracting dimension.

\subsubsection{Failure of PCA}
This example demonstrates how PCA fails with this type of anomaly detection application.  If you try to solve~\ref{eqn:Lwithin_linear} (retrieving the eigenvectors of the within-class variance $C^\text{within}$), the result is $[e_1, e_2, e_3]$ with corresponding eigenvalues $[4, 0.9, 0]$.  The descriptive feature is still marked as the worst offender (because it represents the direction with highest within-set variance).  If instead you attempted to preserve feature directions using PCA the standard way, the eigenvectors of $C^{all}$ are the same $[e_1, e_2, e_3]$, with eigenvalues $[80, 1, 0]$.  In this case, you happen to get lucky that the distracting feature represents less variation in the overall dataset, but by removing the directions of lower variance, the null space will be removed as well, which we already marked as important for anomaly detection.  The descriptive-distracting tradeoff must be considered to avoid these two pitfalls.

\begin{figure}
\centering
  \centering
  \includegraphics[width=.6\linewidth]{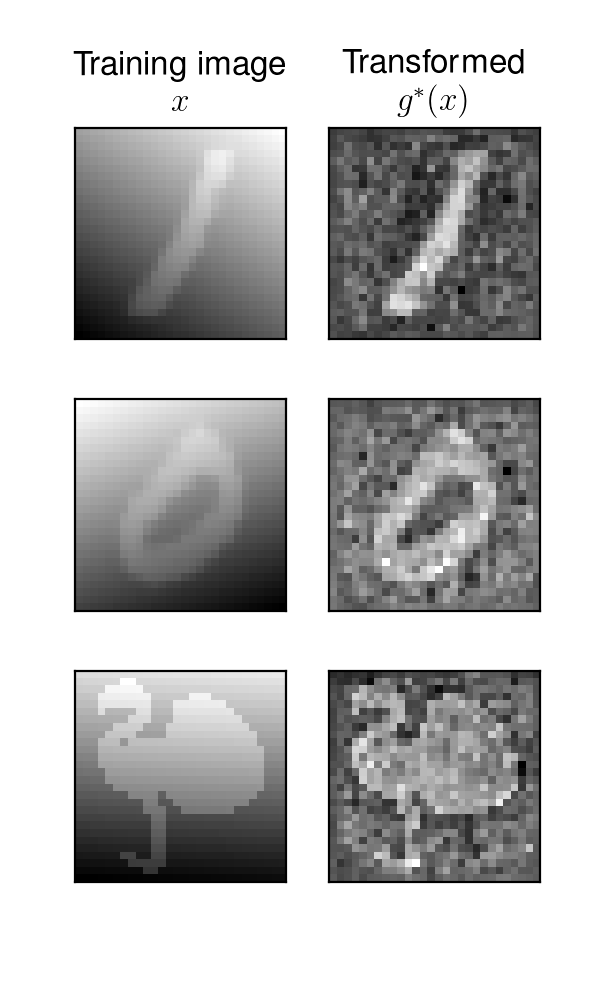}
  \caption{Examples of synthetic images from the training sets (digits 0 and 1 from EMNIST; flamingo from Caltech101 Silhouttes.  The transformed image is projected back into pixel space for visualization purposes (even though $g^*(x)$ has $692$ dimensions instead of $784$).}
  \label{fig:training_images}
\end{figure}

\begin{figure*}
\centering
  \centering
  \includegraphics[width=.95\linewidth]{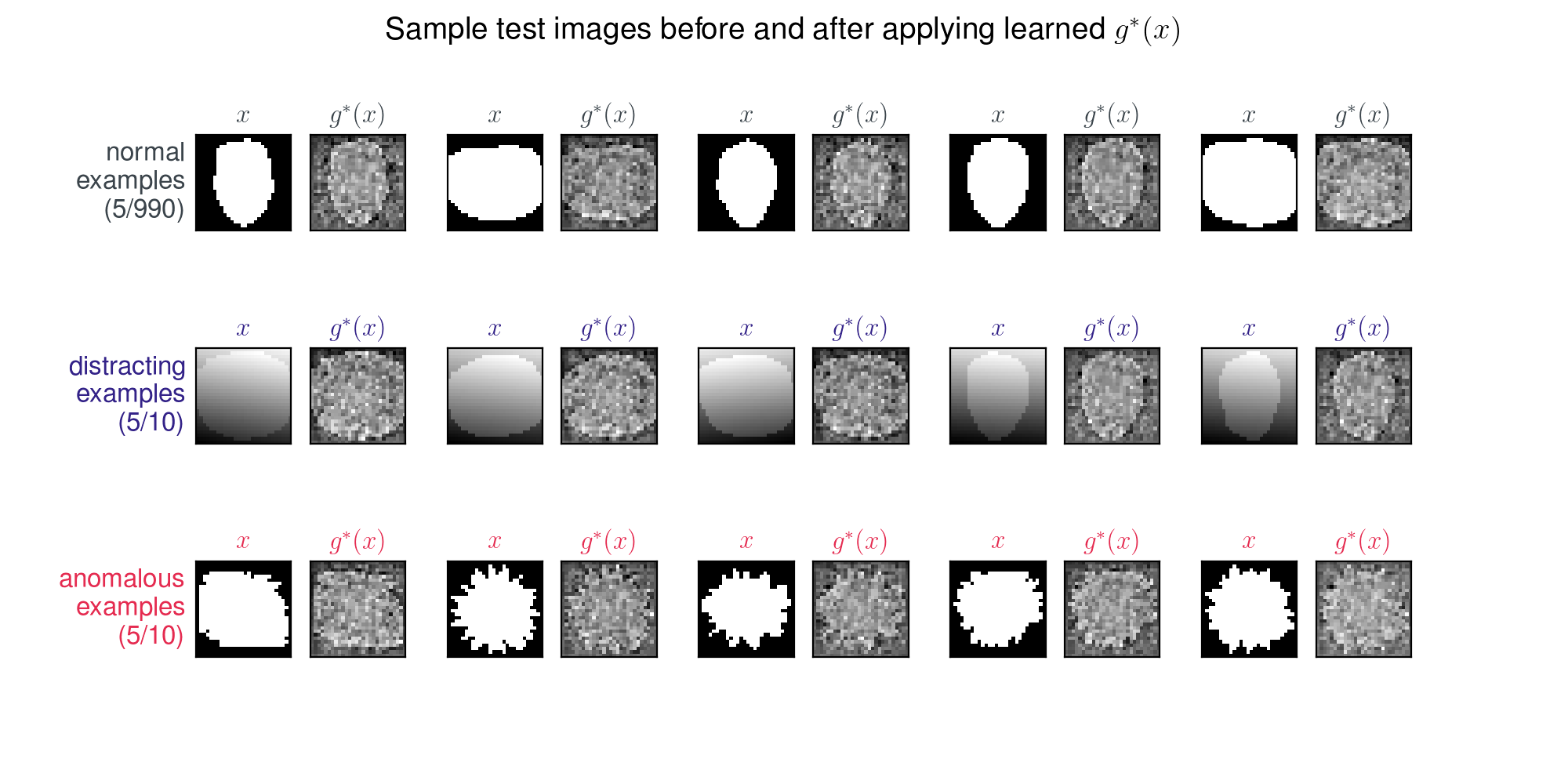}
  \caption{Examples of testing images of each type.  Note that the illumination is removed from the distracting images, while the rest of the descriptive power in the image is left largely intact.}
  \label{fig:testing_images}
\end{figure*}

\subsection{Illustrative example}
\begin{figure*}[t]
\centering
\begin{subfigure}{1\linewidth}
  \centering
  \includegraphics[width=.95\linewidth]{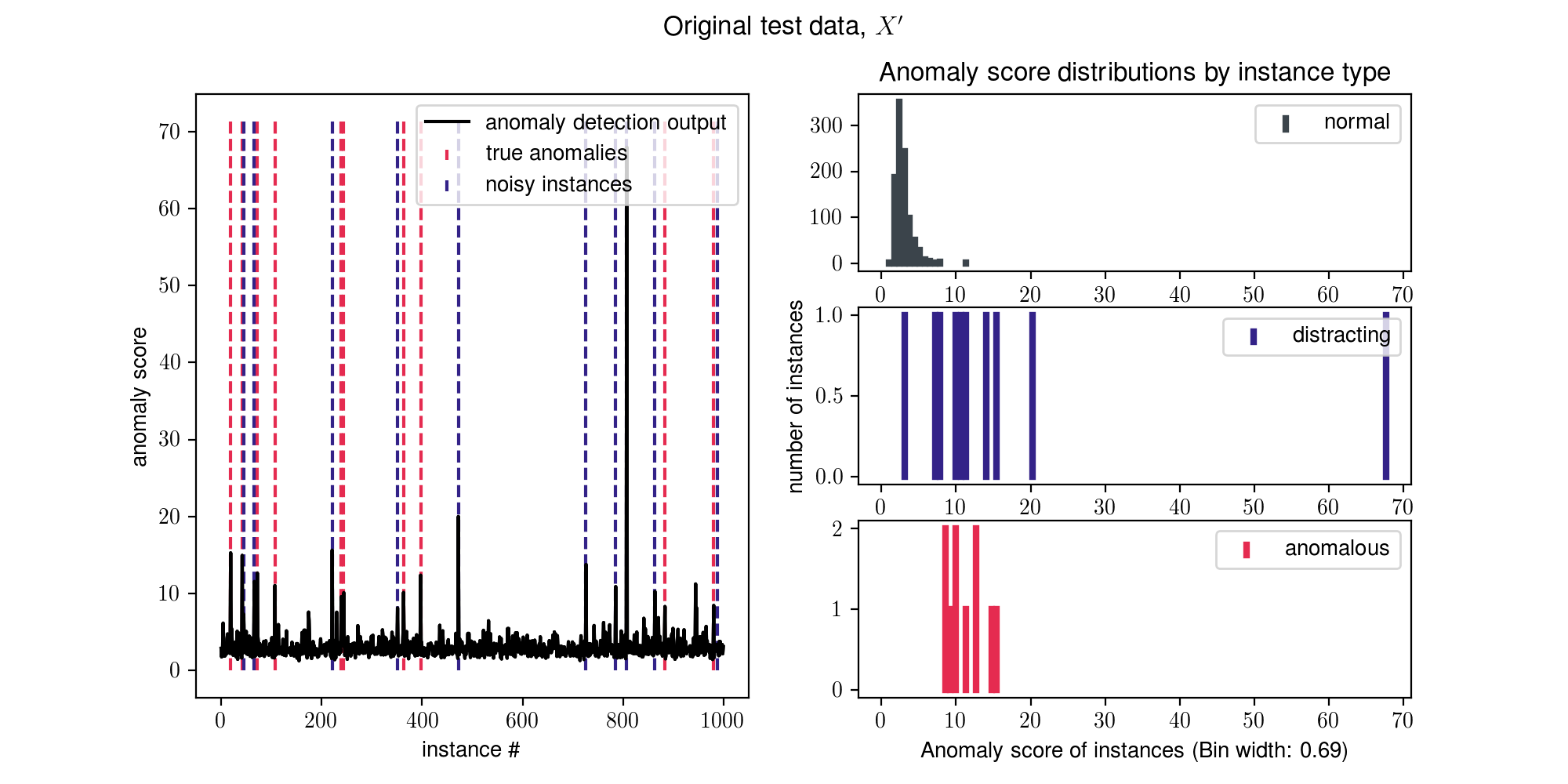}
\end{subfigure}%
\par\medskip  
\begin{subfigure}{1\linewidth}
  \centering
  \includegraphics[width=.95\linewidth]{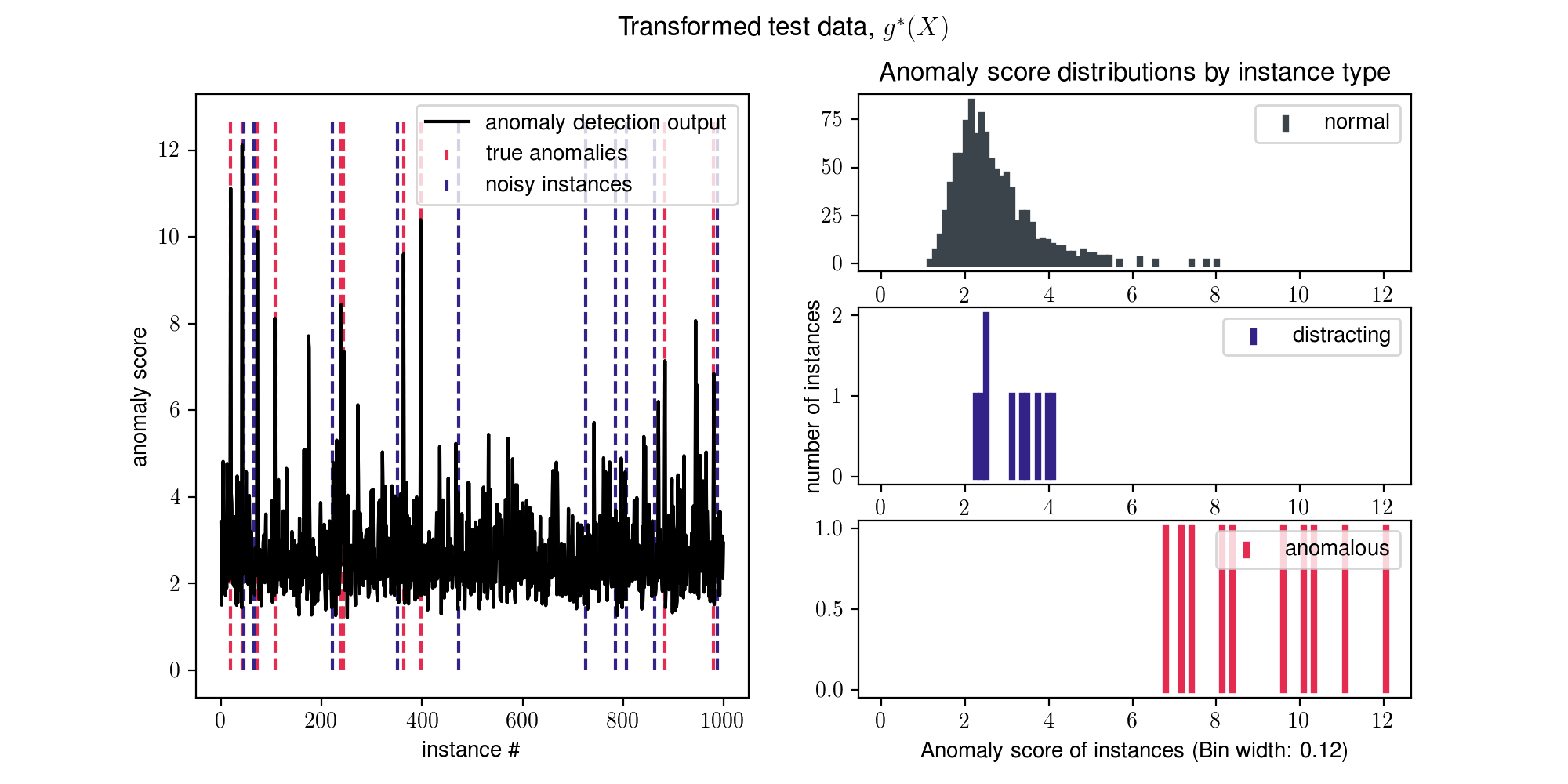}
\end{subfigure}
\caption{(Top.) Before learning the distracting features, even a scale-invariant, discriminative anomaly detection algorithm gives high anomaly scores to the examples with synthetic illumination, largely missing the true anomalies. (Bottom.) After learning which directions correspond to distracting features, the anomaly detection algorithm is able to detect more true anomalies and produce fewer false positives.}
\label{fig:anomaly_original_and_transformed}
\end{figure*}

To demonstrate the method on a familiar dataset, we created a synthetic modification of the EMNIST dataset (handwritten numbers and letters) and Caltech 101 Silhouttes dataset (ground truth masks shrunk to the characteristic 28x28 MNIST size). The full set of EMNIST characters are chosen for training in addition to all but two of the silhoutte classes (which are used for testing).  100 images from each class are chosen as different sets.  Illumination gradients were added to a subset of the images with randomly distributed amplitudes (normal distribution) and angles (uniform distribution).  Example of synthetically modified images are shown in (Figure~\ref{fig:training_images}).  In order to sufficiently represent the descriptive space, we rotated and flipped each of the classes to create more training sets, for a total of $954$.  The test set is constructed from the remaining two characters in the Caltech101 silhouttes dataset (\textit{Faces 2} and \textit{Sunflowers}).  At first, the anomaly detection algorithm (~\cite{delgiorno2016discriminative}) is unable to successfully find the anomalies (sunflowers) hidden among the face silhouttes in the dataset.  A set of training sets of different digits are given to the informative features algorithm and the illumination is correctly identified among the eigenvectors with the highest values.  Choosing a cutoff of $.999$ for the eigenvalues, 92 eigenvectors were removed, leaving a 692 dimensional linear projection, $g^*(x)$.  When these this transformation is applied (Figure~\ref{fig:testing_images}), an anomaly detection algorithm is able to correctly identify the anomalies and is better at ignoring the distracting illuminated normal instances (Figure~\ref{fig:anomaly_original_and_transformed}).


\section{Discussion}
The FOCUS feature selection method can aid existing anomaly detection algorithms by identifying and extracting distracting features from only sets of normal data.  It is simple and easy to implement as well as efficient to train.  Because the covariance matrices can be computed in a streaming fashion, the memory requirements are small, and the computational complexity is dominated by the computation of the covariance matrices, $O(Mnd^2)$, where $n$ is the number of data points, $M$ is the number of training sets, and $d$ is the number of features.

This method is uniquely suited for the anomaly detection setting, and avoids the pitfalls of other methods like PCA, which are unable to evaluate the tradeoff between description and distraction.  Because FOCUS only uses normal data and requires no other class labels (beyond set assignments), the method won't bias the feature set toward notions of anomalies, and can constantly be tuned by adding more normal sets of data.  However, if the number of features is large, a large number of training sets are required if preserving most of the feature space is desired (in general, at least as large as the number of feature dimensions).  This is still suitable for many scenarios when offline data discovery and training is low-cost and worth it for an optimal linear mapping that removes false positives in test sets of different contexts.

\section{Extensions}
\subsection{Human in-the-loop training}
We assumed that training sets were provided before test time and the informative features algorithm occurs before test time.  However, in practical settings, anomaly detection algorithms often flag human users with data with the expectation that humans will post-process it.  We can leverage this setting to ask humans to feed back examples of incorrect detections; this would explicitly penalize parts of the feature space the anomaly detection was incorrectly focused on.  This online learning setting would behave better in practical scenarios and allow the algorithms to acquire more training data for future test sets.
\subsection{Regularization for anomaly detection systems}
The current proposed method involved preprocessing the data before running it through anomaly detection.  One could imagine instead carrying around the objective with the training data as a regularizer as the anomaly detection algorithm runs at test time.  This would penalize models or classifiers that would perform incorrectly on the training data.  This regularization would likely lead to different results (especially in the nonlinear case) when it is used as a regularizer rather than a preprocessing step.
\subsection{Nonlinear generalization}
Kernelizing the objective function would allow the algorithm to learn nonlinear objectives, though at the cost of speed and memory.  It would nonetheless be interesting to see how to include this objective in more complex architectures, including neural networks. 
\subsection{Other applications}
Looking at the space of machine learning problems more generally, this method's key idea is to use \textit{sets} rather than \textit{classes} of data, in the sense that the sets of data have little to do with the task at test time.  One obvious application of this method would be for pure denoising.  It would also be interesting to explicitly teach an algorithm invariances for more standard ML problems; e.g. - for classification when examples of invariances within certain classes are sparse or nonexistent, or it is easy to generate examples of invariances the algorithm must learn in other contexts.

\bibliography{example_paper}
\bibliographystyle{icml2016}

\onecolumn
\newpage

\section{Appendix}

\subsection{Loss Function to Generalized Eigenvalue Problem} \label{section:lossgeneig}

We derive the general case where we specify a prior over the training sets.  The primary reason for this is when the number of samples per set differs, the relative importance of each set needs to be established.  If each training set should be weighted equally, we need to enforce a uniform prior over each training set.  The prior can be thought of either as weighting the sum of squared distances to each point in that set, or as sampling from the 'underlying distribution' of the training sets to balance the sets before computing the sum of squared distances.  Any other weighting scheme can be chosen over the training sets by adjusting the prior (weighting) distribution $P_m$.

Let $X,Y$ be discrete independent variables representing samples from the training set.  Each are dependent on the prior over the training sets.  The conditional distribution of $X|m$ is uniform over each point in set $m$.  The prior $P_m$ can then be chosen based on whether each set should have weights that are uniform ($P_m=\frac{1}{M}$), proportional to the size of the set ($P_m=\frac{n_m}{n_\text{all}}$), or another discrete distribution that represents a notion of relative set importance.

\subsubsection{Fidelity term: $L^\text{all}$} \label{section:Call_linear}

\begin{subequations}
\begin{empheq}{align}
    L^{\text{all}}_w &=\left[\E_X\E_Y (w^TX-w^TY)(w^TX-w^TY)^T\right]\\
    &=w^T\left[\E_X\E_Y (X-Y)(X-Y)^T\right]w\\
    &=w^T\left[\E_X\E_Y (X-c+c-Y)(X-c+c-Y)^T\right]w\\
    &=w^T\left[\E_X\E_Y (X-c)(X-c)^T + (X-c)(c-Y)^T + (c-Y)(X-c)^T + (Y-c)(Y-c)^T\right]w\\
    &=w^T\left[2\E_X\E_Y\left[(X-c)(X-c)^T\right]
    + \E_X\E_Y\left[(X-c)(c-Y)^T\right]
    + \E_X\E_Y \left[(c-Y)(X-c)^T\right] \right]w\\
    &=w^T\left[2\E_X\E_Y\left[(X-c)(X-c)^T\right]
    + \E_X\left[(X-c)\E_Y(c-Y)^T\right]
    + \E_Y \left[(c-Y)\E_X(X-c)^T\right] \right]w\\
    & \text{Choose } c=\muhat_\text{all} \text{ such that } \E_X \muhat_\text{all} =\E_X X\\
    &=w^T\left[2\E_X\E_Y\left[(X-\muhat_\text{all})(X-\muhat_\text{all})^T\right] 
    + \E_X\left[(X-\muhat_\text{all})\E_Y(\muhat_\text{all}-Y)^T\right] 
    + \E_Y \left[(\muhat_\text{all}-Y)\E_X(X-\muhat_\text{all})^T\right]\right]w\\
    &=w^T\left[2\E_X\E_Y\left[(X-\muhat_\text{all})(X-\muhat_\text{all})^T\right] 
    + \E_X\left[(X-\muhat_\text{all})(0)\right] 
    + \E_Y \left[(\muhat_\text{all}-Y)(0)\right]\right]w\\
    &=w^T\left[2\E_X\left[(X-\muhat_\text{all})(X-\muhat_\text{all})^T\right]\right]w\\
    &=w^T\left[2\E_m\E_{X|m}\left[(X-\muhat_\text{all})(X-\muhat_\text{all})^T\right]\right]w\\
    &=2w^TC^\text{all} w\\
\end{empheq}
\end{subequations}

Where $C^\text{all} = \E_m\E_{X|m}\left[(X-\muhat_\text{all})(X-\muhat_\text{all})^T\right]$.

The empirical estimate is then computed as follows:

\begin{empheq}{align}
    \hat{C}^\text{all}&=\E_m\E_{X|m}\left[(X-\muhat_\text{all})(X-\muhat_\text{all})^T\right]\\
    &=\sum_{m=1}^M P_m \sum_{x\in \mathcal{X}^m} P_{x|m} (X-\muhat_\text{all})(X-\muhat_\text{all})^T \\
    &=  \sum_{m=1}^M \frac{1}{n_m} P_m \sum_{x\in \mathcal{X}^m}(X-\muhat_\text{all})(X-\muhat_\text{all})^T\\
\end{empheq}

\begin{empheq}{align}
    &\muhat_\text{all}=\sum_m P_m \muhat_m\\
    &\muhat_m=\frac{1}{n_m}\sum_{x \in \mathcal{X}^m}x
\end{empheq}

If we wish to weight each training set equally (uniform prior over sets: $P_m=\frac{1}{M}$), this becomes:
\begin{empheq}{align}
    \hat{C}^\text{all} &= \frac{1}{M} \sum_{m=1}^M \frac{1}{n_m} \sum_{x\in \mathcal{X}^m}(x-\muhat_\text{all})(x-\muhat_\text{all})^T\\
    &\muhat_\text{all}= \frac{1}{M} \sum_m \frac{1}{n_m}\sum_{x \in \mathcal{X}^m}x\\
\end{empheq}

If we wish to weight each training set according to the number of points in it (uniform prior over datapoints: $P_m=\frac{n_m}{n_\text{all}}$), this becomes:
\begin{empheq}{align}
    \hat{C}^\text{all} &= \frac{1}{n_\text{all}} \sum_{m=1}^M \sum_{x\in \mathcal{X}^m}(x-\muhat_\text{all})(x-\muhat_\text{all})^T\\
    &\muhat_\text{all} = \frac{1}{n_\text{all}} \sum_m \sum_{x \in \mathcal{X}^m} x
\end{empheq}

Note that when the training sets are balanced ($\frac{1}{n_m}$ is the same for all $m$), these estimators are the same.

\subsubsection{Invariance term: $L^\text{within}$} \label{section:Cwithin_linear}

\begin{subequations}
\begin{empheq}{align}
    L^{\text{within}}_w &=\left[\E_m\E_{X|m}\E_{Y|m} (w^TX-w^TY)(w^TX-w^TY)^T\right]\\
    &=w^T\left[     \E_m\E_{X|m}\E_{Y|m} \left[(X-c)(X-c)^T
                +   (X-c)(c-Y)^T
                +   (c-Y)(X-c)^T
                +   (Y-c)(Y-c)^T\right]
    \right]w\\
    &=w^T\left[     2\E_m\E_{X|m}\E_{Y|m} (X-c)(X-c)^T
                +   2\E_m\E_{X|m}\E_{Y|m} (c-X)(Y-c)^T
    \right]w\\
    &=w^T\left[     2\E_m\E_{X|m}\E_{Y|m} (X-c)(X-c)^T
                +   2\E_m\E_{X|m}(c-X)\E_{Y|m}(Y-c)^T
    \right]w\\
    & \text{Choose } c=\muhat_m \text{ such that } \E_{X|m} \muhat_m =\E_{X|m} X\\
    &=w^T\left[     2\E_m\E_{X|m}\E_{Y|m} (X-\muhat_m)(X-\muhat_m)^T
                +   2\E_m\E_{X|m}(\muhat_m-X)\E_{Y|m}(Y-\muhat_m)^T
    \right]w\\
    &=w^T\left[     2\E_m\E_{X|m}\E_{Y|m} (X-\muhat_m)(X-\muhat_m)^T
                +   2\E_m\E_{X|m}(\muhat_m-X)(0)
    \right]w\\
    &=w^T\left[     2\E_m\E_{X|m}\E_{Y|m} (X-\muhat_m)(X-\muhat_m)^T
    \right]w\\
    &=2 w^T C^\text{within} w\\
\end{empheq}
\end{subequations}

Our estimators are then computed as follows:

\begin{empheq}{align}
    \hat{C}^\text{within}&=\E_m\E_{X|m}\left[(X-\muhat_m)(X-\muhat_m)^T\right]\\
    &=\sum_{m=1}^M P_m \sum_{x\in \mathcal{X}^m} P_{x|m} (X-\muhat_m)(X-\muhat_m)^T \\
    &=\sum_{m=1}^M \frac{1}{n_m} P_m \sum_{x\in \mathcal{X}^m}(x-\muhat_m)(x-\muhat_m)^T\\
\end{empheq}

\begin{empheq}{align}
    \muhat_m&=\frac{1}{n_m}\sum_{x \in \mathcal{X}^m}x
\end{empheq}

If we wish to weight each training set equally (uniform prior), this becomes:
\begin{empheq}{align}
    \hat{C}^\text{within} &= \frac{1}{M} \sum_{m=1}^M \frac{1}{n_m} \sum_{x\in \mathcal{X}^m}(x-\muhat_m)(x-\muhat_m)^T\\
\end{empheq}

Note that when the training sets are balanced ($\frac{1}{n_m}$ is the same for all $m$), these estimators are the same.

\subsubsection{Loss to generalized eigenvalue problem} \label{section:geneig}
Maximizing $L^{\text{within}}$ and minimizing $L^{\text{all}}$ can be performed using the standard derivation of the generalized eigenvalue problem:
\begin{empheq}{align}
    &\arg \max_{g,\lambda} L_g^{\text{within}}(\mathcal{X})-\lambda L_g^{\text{all}}(\mathcal{X}) \\
    = &\arg\max_w \frac{2n_\text{all}w^T C^\text{within} w}{2n_\text{all}w^T C^\text{all} w}\\
    = &\frac{w^T C^\text{within} w}{w^T C^\text{all} w}
\end{empheq}

\subsection{Relation between $C^\text{all}$ and $C^\text{within}$: $Q$} \label{section:within_between_nullspace_bootstrap}

\textbf{Theorem:}
\begin{empheq}{align}
    C^\text{all} = C^\text{within} + Q; \qquad Q=\E_m (\muhat_m \muhat_m^T) - \muhat_\text{all}\muhat_\text{all}^T
\end{empheq}

$Q$ has at most rank $M-1$ (the rank is reduced by one for each repeated mean among the training sets).  

\textbf{Corollaries:}
\begin{empheq}{align}
    \text{1. }&\text{nullspace}(C^\text{all}) \subseteq \text{nullspace}(C^\text{within})\\
    \text{2. }&\text{span}(C^\text{all}) \supseteq \text{span}(C^\text{within})\\
    \text{3. }&\text{rank}(C^\text{all}) \ge \text{rank}(C^\text{within})
\end{empheq}

\textbf{Proof:}\\

\begin{subequations}
\begin{empheq}{align}
    C^\text{within} &=\E_m \E_{X|m} (X-\muhat_m)(X-\muhat_m)^T\\
                    &=\E_m \E_{X|m} (X-\muhat_\text{all}+\muhat_\text{all}-\muhat_m)(X-\muhat_\text{all}+\muhat_\text{all}-\muhat_m)^T\\
                    &=\E_m \E_{X|m}[
                    (X-\muhat_\text{all})(X-\muhat_\text{all})^T
                    + (X-\muhat_\text{all})(\muhat_\text{all}-\muhat_m)^T \nonumber \\
                    & \qquad + (\muhat_\text{all}-\muhat_m)(X-\muhat_\text{all})^T
                    + (\muhat_\text{all}-\muhat_m)(\muhat_\text{all}-\muhat_m)^T
                    ]\\
                    &=C^\text{all}
                    + \E_m (\E_{X|m}[X]-\muhat_\text{all})(\muhat_\text{all}-\muhat_m)^T
                    + \E_m (\muhat_\text{all}-\muhat_m)(\E_{X|m}[X]-\muhat_\text{all})^T
                    + \E_m (\muhat_\text{all}-\muhat_m)(\muhat_\text{all}-\muhat_m)^T\\
                &\text{Recall } \E[\muhat_m]=\E_{X|m}[X]\\
                    &=C^\text{all}
                    - \E_m (\muhat_\text{all}-\muhat_m)(\muhat_\text{all}-\muhat_m)^T
                    - \E_m (\muhat_\text{all}-\muhat_m)(\muhat_\text{all}-\muhat_m)^T
                    + \E_m (\muhat_\text{all}-\muhat_m)(\muhat_\text{all}-\muhat_m)^T\\
                    &=C^\text{all} - Q\\
\end{empheq}
\end{subequations}

\begin{subequations}
\begin{empheq}{align}
                    Q&=\E_m[(\muhat_\text{all} - \muhat_m)(\muhat_\text{all} - \muhat_m)^T\\
                    &=\muhat_\text{all}\muhat_\text{all}^T
                    - \muhat_\text{all} \left[\E_m \muhat_m^T\right]
                    - \left[\E_m \muhat_m\right] \muhat_\text{all}^T
                    + \E_m \muhat_m \muhat_m^T\\
                &\text{Recall } \E[\muhat_\text{all}]=\E_m\E_{X|m}X=\E_m\muhat_m\\
                    &=\E_m (\muhat_m \muhat_m^T) - \muhat_\text{all}\muhat_\text{all}^T
\end{empheq}
\end{subequations}

$\hat{Q}$ can be computed as follows:

\begin{empheq}{align}
    \hat{Q}=\sum_{m=1}^M (P_m \muhat_m \muhat_m^T) - \muhat_\text{all}\muhat_\text{all}^T
\end{empheq}

\subsection{Useful Properties Relating to Covariance Matrices}
\begin{lemma} \label{lemma:CovComp}
\begin{empheq}{align}
    \sum_{x,y \in \mathcal{X}}(x-y)(x-y)^T=2n\sum_{x \in \mathcal{X}}(x-\mu)(x-\mu)^T
    \label{eqn:CovComp}
\end{empheq}
\end{lemma}

\textbf{Proof:}\\
\begin{subequations}
\begin{empheq}{align}
    &\sum_{x,y \in \mathcal{X}}(x-y)(x-y)^T\\
    =&\sum_{x,y \in \mathcal{X}} \left((x - \mu) + (\mu - y)\right)\left((x - \mu) + (\mu - y)\right)^T \\
    =&\sum_{x,y \in \mathcal{X}} (x - \mu)(x - \mu)^T + (y - \mu) (y - \mu)^T \\
        &\qquad -(x - \mu)(y - \mu)^T - (y - \mu)(x - \mu)^T \\
    =& \sum_{y \in \mathcal{X}}\left[\sum_{y \in \mathcal{X}} (x - \mu)(x - \mu)^T\right] + \sum_{x \in \mathcal{X}}\left[\sum_{y \in \mathcal{X}}(y - \mu)(y - \mu)^T\right]\\
        &\qquad -\sum_{x \in \mathcal{X}}\sum_{y \in \mathcal{X}}\left[(x - \mu)(y - \mu)^T - (y - \mu)(x - \mu)^T\right] \\
    =& n\left[\sum_{x \in \mathcal{X}} \left(x - \mu\right)(x - \mu)^T\right] + n\left[\sum_{y \in \mathcal{X}}(y - \mu)(y - \mu)^T\right]  \nonumber\\
        &\qquad -\sum_{x \in \mathcal{X}}(x - \mu)\left(\sum_{y \in \mathcal{X}}(y - \mu)^T\right) - \sum_{y \in \mathcal{X}}(y - \mu)\left(\sum_{x \in \mathcal{X}}(x - \mu)^T\right) \\
    =& 2n\left[\sum_{x \in \mathcal{X}} (x - \mu)(x - \mu)^T\right]  \nonumber
        -\sum_{x \in \mathcal{X}}(x - \mu)(0) - \sum_{x \in \mathcal{X}}(x - \mu)(0) \\
    =& 2n\left[\sum_{x \in \mathcal{X}} (x - \mu)(x - \mu)^T\right]
\end{empheq}
\end{subequations}

\begin{lemma} \label{lemma:CovPlusOuterProdMean}
\begin{subequations}
\begin{empheq}{align}
    &\sum_{x \in \mathcal{X}}(x-\mu)(x-\mu)^T=\sum_{x \in \mathcal{X}}xx^T - n\mu\mu^T \label{eqn:CovPlusOuterProdMean}\\
    &\text{equivalently,}  \nonumber\\
    &(X-\mu)^T(X-\mu)=X^TX - n\mu\mu^T \label{eqn:matrices_CovPlusOuterProdMean}
\end{empheq}
\end{subequations}
\end{lemma}

\end{document}